\documentclass[conference]{IEEEtran}
\IEEEoverridecommandlockouts
\usepackage{cite}
\usepackage{amsmath,amssymb,amsfonts}
 \usepackage{verbatim}
\usepackage{graphicx}
 \usepackage{url} 
\usepackage{mdframed}
\usepackage{multirow}
\usepackage{booktabs}
\usepackage[table,xcdraw]{xcolor}
\usepackage{textcomp}
\usepackage{xcolor}
\usepackage{algpseudocode}
\usepackage{algorithm}
\usepackage[table]{xcolor} 
\usepackage{multirow}      
\usepackage{booktabs}      
\usepackage[utf8]{inputenc}
\usepackage{tcolorbox}
\usepackage{subcaption}
\usepackage[font=small]{caption}
\tcbuselibrary{skins}

\def\BibTeX{{\rm B\kern-.05em{\sc i\kern-.025em b}\kern-.08em
    T\kern-.1667em\lower.7ex\hbox{E}\kern-.125emX}}
\begin{document}

\title{Understanding the Energy Scaling of Large Language Model Inference Across Context Lengths and Attention Architectures}


\author{
\IEEEauthorblockN{Molka Chkir\IEEEauthorrefmark{1}, Syed Muhammad Danish\IEEEauthorrefmark{1}, Jos H\"oll\IEEEauthorrefmark{2}, Arghavan Asad\IEEEauthorrefmark{1}} 
\IEEEauthorblockA{\IEEEauthorrefmark{1}Algoma University, Brampton, Canada \\
\IEEEauthorblockA{\IEEEauthorrefmark{2}School of Informatics, Reutlingen University, Reutlingen}
Emails: \{mchkir, syed.danish, arghavan.asad\}@algomau.ca, jos.hoell@reutlingen-university.de }
}

\maketitle

\begin{abstract}

The growing adoption of large language models (LLMs) has raised increasing concerns about the energy consumption and environmental impact of inference. This paper presents a systematic empirical study of decode-phase energy consumption across representative open-source LLMs employing Multi-Head Attention (MHA), Grouped Query Attention (GQA), and Grouped Query Attention with Sliding Window Attention (SWA) to characterize how attention architecture influences decode-phase energy consumption under varying inference workloads. We evaluate four models across different context lengths, batch sizes, and generation workloads while measuring GPU energy using NVIDIA hardware counters. We examine the effects of context length, attention mechanism, Key-Value (KV) cache growth, and batching on decode-phase energy consumption. Results show that attention mechanism is the primary factor governing how decode energy scales with context length. MHA models exhibit substantially steeper energy growth than GQA models, whereas GQA with SWA maintains nearly constant energy consumption. We further show that model size primarily determines absolute energy consumption, while batching reduces both energy per generated token and request latency by up to 87\%. These findings provide practical guidance for selecting energy-efficient LLM architectures and inference configurations.

\end{abstract}


\section{Introduction}
LLMs \cite{rahman2026refactorcoderqa} have become the foundation of modern AI applications, enabling capabilities such as conversational assistants, code generation \cite{ashraf2025toward}, question answering, and summarization \cite{brown2020language}. As these models transition from research prototypes to production services, their environmental impact has become an increasing concern. The computational demands of training and serving LLMs result in substantial energy consumption, carbon emissions, and water usage \cite{luccioni2024power}. While model training has traditionally received the most attention, recent studies estimate that inference may account for up to 90\% of a model's total lifecycle energy consumption \cite{wu2022sustainable}. Consequently, improving inference efficiency has become a key challenge for the sustainable deployment of LLM-powered applications.

Despite growing interest in sustainable LLM inference, existing work remains limited in scope. Most prior studies have focused on the environmental impact of model training \cite{luccioni2024power,patterson2021carbon} or have evaluated inference energy using only a small number of models under fixed workloads \cite{ali2026assessing}, providing limited insight into the factors that govern inference energy consumption. One particularly important but underexplored factor is the evolution of attention mechanisms across modern LLM architectures. Contemporary models employ increasingly diverse attention designs, including MHA \cite{vaswani2017attention}, GQA \cite{ainslie2023gqa}, and Grouped Query Attention combined with SWA \cite{beltagy2020longformer}. These mechanisms differ fundamentally in how Key-Value states are stored and accessed during autoregressive decoding, leading to different computational and memory behaviors as context length increases. Although these architectural innovations were introduced primarily to improve inference efficiency, their impact on decode-phase energy scaling has not been systematically characterized. Consequently, developers lack evidence-based guidance on how attention architecture influences inference energy and the associated deployment trade-offs.

To address this gap, we present a systematic empirical study of decode-phase energy consumption across representative open-source LLMs employing different attention mechanisms. Rather than measuring only overall energy consumption, our objective is to understand how decode energy scales under realistic inference workloads. Specifically, we investigate how context length, attention mechanism design, KV cache growth during autoregressive generation, and request batching influence decode-phase energy consumption. We address the following research questions (RQ):

\noindent -- \textbf{RQ1:} How does decode energy per generated token scale with increasing context length?

\noindent -- \textbf{RQ2:} How do different attention mechanisms influence the scaling of decode energy with context length?

\noindent -- \textbf{RQ3:} How does token position within a generation sequence affect decode energy as the KV cache grows?

\noindent -- \textbf{RQ4:} How does batching affect energy per generated token and latency across models and context lengths?

To answer these research questions, we conduct a comprehensive empirical evaluation of four representative open-source LLMs spanning three attention mechanisms: OPT-1.3B and Phi-3 Mini (MHA), Gemma-2-2B (GQA), and Mistral-7B (GQA with SWA). Our results show that attention mechanism design is the primary factor governing decode-energy scaling. MHA models exhibit substantially steeper energy growth than GQA models, whereas the combination of GQA and SWA maintains nearly constant energy consumption as context length increases. Within a generation sequence, decode energy progressively increases for MHA models as the KV cache grows, remains largely stable for GQA models, and is effectively constant with SWA. In contrast, although attention architecture determines energy scaling behavior, absolute energy consumption is primarily driven by model size, meaning that a smaller MHA model may consume less energy per generated token than a larger GQA model despite its less efficient attention mechanism. Finally, batching consistently improves inference efficiency, reducing both energy per generated token and request latency by up to 87\% across all evaluated models and context lengths. Together, these findings provide a systematic characterization of how modern attention mechanisms influence decode-phase energy scaling and offer practical guidance for energy-aware deployment of large language models.

The rest of the paper is organized as follows. Section II presents the related work. Section III presents the overall methodology and experimental setup. Section IV discusses the results and analysis for each research question. Section V presents the limitations, while Section VI concludes the paper.

\section{Related Work}
Early studies on the environmental impact of LLMs primarily focused on the training phase. Patterson et al. \cite{patterson2021carbon} quantified the carbon emissions associated with training large neural networks, highlighting the substantial environmental cost of developing frontier models. Extending this perspective, Wu et al. \cite{wu2022sustainable} analyzed the complete lifecycle of AI systems and estimated that inference may account for up to 90\% of total energy consumption once models are deployed at scale. These studies established the importance of sustainable AI but provided limited insight into the factors governing inference-time energy consumption.

As inference emerged as the dominant source of operational energy use, several studies began characterizing the energy efficiency of LLM inference. Samsi et al. \cite{samsi2023words} conducted one of the earliest benchmarking studies of LLM inference, evaluating multiple LLaMA models on NVIDIA V100 and A100 GPUs across question-answering and mathematical reasoning tasks. Luccioni et al. \cite{luccioni2024power} systematically compared inference costs across diverse machine learning models, showing that generative models consume substantially more energy than task-specific systems and that energy consumption increases with model complexity. More recently, Husom et al. \cite{husom2024price} demonstrated that prompt length significantly influences inference energy, while Stojkovic et al. \cite{stojkovic2024greener} investigated system-level strategies for improving the energy efficiency of LLM inference. Similarly, TokenPowerBench \cite{tokenpowerbench2024} analyzed per-token power consumption under different serving configurations, and SustainableNLP \cite{sustainablenlp2025} benchmarked inference energy across a diverse collection of models and tasks, highlighting the influence of model architecture and workload characteristics. Although these studies significantly advanced the understanding of inference energy, they primarily reported aggregate measurements and did not isolate the decode phase or systematically investigate how attention mechanism design affects energy scaling as context length increases.

In parallel, substantial research has focused on improving transformer efficiency through architectural innovations. Vaswani et al. \cite{vaswani2017attention} introduced MHA, which forms the foundation of modern transformer architectures. Ainslie et al. \cite{ainslie2023gqa} proposed GQA, reducing decoding cost by allowing multiple query heads to share a smaller set of key-value heads. Beltagy et al. \cite{beltagy2020longformer} introduced SWA, limiting attention to a fixed local context to reduce computational complexity for long sequences. Building on these ideas, Jiang et al. \cite{jiang2023mistral} incorporated GQA and SWA into Mistral-7B, demonstrating substantial improvements in inference throughput for long-context generation. Beyond architectural changes, Pope et al. \cite{pope2023efficiently} investigated transformer inference efficiency from a systems perspective, while Chung et al. \cite{chung2025joules} studied quantization, batching, and serving strategies, showing that batch size plays a critical role in reducing energy per generated token. Despite these advances, existing work has largely focused on improving computational efficiency rather than understanding how different attention mechanisms fundamentally influence decode-phase energy consumption.
\subsection{Novelty of this work}

Existing studies have investigated inference-time energy consumption from perspectives such as model benchmarking, hardware platforms, prompt characteristics, and serving strategies. However, none have systematically characterized how attention mechanism design influences decode-phase energy scaling as context length increases. Previous work either reports aggregate inference energy without isolating the decode phase, focuses on a single model family, or evaluates fixed workloads without jointly considering context length, token generation position, and batching. In contrast, this work specifically isolates the decode phase from the prefill phase and systematically characterizes how context length, attention mechanism, token generation position, and batch size jointly influence decode-phase energy consumption. We evaluate representative LLMs employing MHA, GQA, and GQA with SWA under identical hardware, precision, and measurement conditions using hardware-level GPU energy counters. To the best of our knowledge, this is the first study to directly compare decode-energy scaling across these fundamentally different attention mechanisms.

Our findings provide practical guidance for selecting energy-efficient LLM architectures and inference configurations for deployment. They also provide quantitative evidence of how modern attention mechanisms affect decode-phase energy scaling, informing the design of future energy-efficient LLM architectures.

\section{Methodology}
\begin{figure*}[t]
    \centering
    \includegraphics[width=0.8\linewidth]{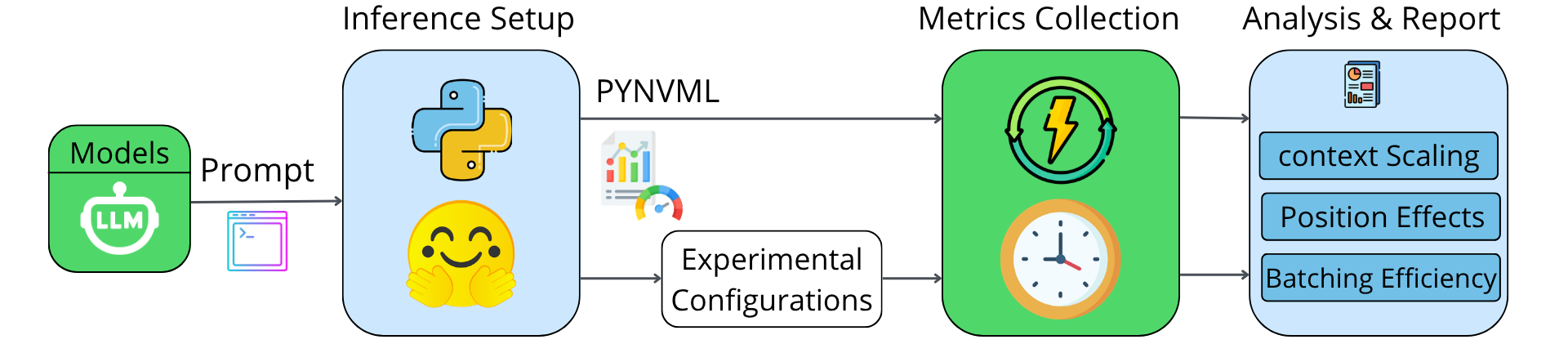}
    \caption{Overall methodology of the proposed work.}
    \label{fig:methodology}
\end{figure*}

This section presents the overall methodology of the proposed study, as illustrated in Fig.~\ref{fig:methodology}. The following subsections describe each component of the methodology in detail.

\subsection{Selection of LLMs}
In this study, we evaluate four representative open-source language models: OPT-1.3B \cite{zhang2022opt}, Gemma-2-2B \cite{gemma2024}, Phi-3 Mini \cite{phi3mini2024}, and Mistral-7B \cite{jiang2023mistral}. These models were selected because they represent different attention mechanisms used in modern transformer architectures. Specifically, OPT-1.3B and Phi-3 Mini employ MHA, Gemma-2-2B employs GQA, and Mistral-7B combines GQA with SWA. This selection enables a systematic comparison of how different attention mechanisms influence decode-phase energy consumption and its scaling with increasing context length. To ensure a fair comparison, all models were downloaded from Hugging Face, executed locally using 16-bit floating-point precision (FP16), and evaluated under identical hardware and software conditions. 



\subsection{Sustainability Metrics}

To evaluate the energy efficiency and performance of decode-phase inference, we consider three metrics: energy per generated token, latency per token, and latency per request.

\noindent\textbf{1) Energy per Generated Token:} This metric represents the average GPU energy required to generate one output token during the decode phase. It is calculated by dividing the total decode energy by the number of generated tokens. This metric enables fair comparisons across models and context lengths and is reported in Joules (J).

\noindent\textbf{2) Latency per Token:} This metric represents the average time required to generate one output token during the decode phase. It is calculated by dividing the total decode time by the number of generated tokens. Together with energy per generated token, it helps distinguish whether energy differences arise from higher power consumption or slower token generation. Latency is reported in milliseconds (ms).

\noindent\textbf{3) Latency per Request:} This metric represents the average time required to process a single request during batched inference. It is calculated by dividing the total decode time by the batch size. This metric captures the effect of batching on inference efficiency and is reported in milliseconds (ms).

\subsection{Inference Setup}
Following model selection, all four models were downloaded from Hugging Face and executed locally using 16-bit floating-point precision (FP16). To ensure consistent inputs across all models and experimental configurations, we constructed a fixed base prompt from a paragraph of general English text and truncated it to the target context length using each model's tokenizer, ensuring that the tokenized input matched the desired context length. Inference was divided into two phases: prefill and decode. During the prefill phase, the input prompt was processed and the Key-Value (KV) cache was populated; this phase was excluded from energy measurements. During the decode phase, the model generated output tokens autoregressively by reusing the KV cache, and only this phase was included in the evaluation. All experiments used greedy decoding, where the highest-probability token was selected at each decoding step to ensure deterministic and reproducible outputs. Before each measurement, two warm-up inference runs were performed to stabilize GPU clock frequencies and thermal conditions.

\subsection{Measurement Environment}
All experiments were conducted on the Narval high-performance computing cluster operated by the Digital Research Alliance of Canada using a dedicated NVIDIA A100-SXM4-40GB GPU. Each experiment was submitted through the SLURM workload manager and executed on a dedicated GPU node to eliminate interference from other workloads. The software environment consisted of Python 3.10, PyTorch with CUDA support, Hugging Face Transformers (v5.3.0), and the NVIDIA Management Library (NVML) accessed through the PyNVML Python interface for hardware-level energy measurement. Each experimental configuration was repeated 10 times to improve measurement reliability. In addition, a 20 ms settling period was applied before and after each energy measurement to allow the GPU power state to stabilize. The complete experimental configuration is summarized in Table~\ref{tab:config}.
\begin{table}[H]
\centering
\caption{Experimental Configuration}
\label{tab:config}
\begin{tabular}{ll}
\hline
\textbf{Parameter} & \textbf{Value} \\
\hline
\hline
GPU & NVIDIA A100-SXM4-40GB \\
Cluster & Narval (Digital Research Alliance of Canada) \\
Job scheduler & SLURM \\
Python version & 3.10 \\
Transformers version & 5.3.0 \\
Energy measurement & PyNVML (NVML hardware counters) \\
Model precision & 16-bit floating point (FP16) \\
Decoding strategy & Greedy decoding \\
Number of runs & 10 per configuration \\
Settling period & 20 ms before and after each reading \\
Warmup passes & 2 full forward passes before measurement \\
Generated tokens & 200 tokens (RQ1, RQ2, RQ4) \\
Chunk size & 100 tokens (RQ3) \\
Context lengths & 128, 512, 1024, 1800 tokens \\
Batch sizes & 1, 2, 4, 8 requests \\
\hline
\end{tabular}
\end{table}

\subsection{Measurement and Analysis}
GPU energy consumption was measured using the NVIDIA Management Library (NVML) through the \texttt{nvmlDeviceGetTotalEnergyConsumption()} API, which provides the cumulative GPU energy consumption in millijoules. Energy was recorded immediately before the decode phase and immediately after the final output token was generated. The difference between the two readings, converted to Joules, represents the total decode-phase energy for each run. To ensure reliable measurements, all experiments generated at least 200 output tokens (approximately 2 seconds of decoding), allowing multiple updates of the hardware energy counter during each measurement interval. Each experimental configuration was repeated 10 times. For every configuration, we report the mean energy consumption, while the standard deviation and coefficient of variation were used to assess measurement stability. To compare models of different sizes fairly, energy scaling was analyzed using the slope of the mean energy-per-token curve across increasing context lengths rather than absolute energy values.

\section{Results}
In this section, we present and analyze the evaluation results of various LLMs to address RQ1, RQ2, RQ3 and RQ4.
\begin{figure}[t]
    \centering
    \includegraphics[width=\linewidth]{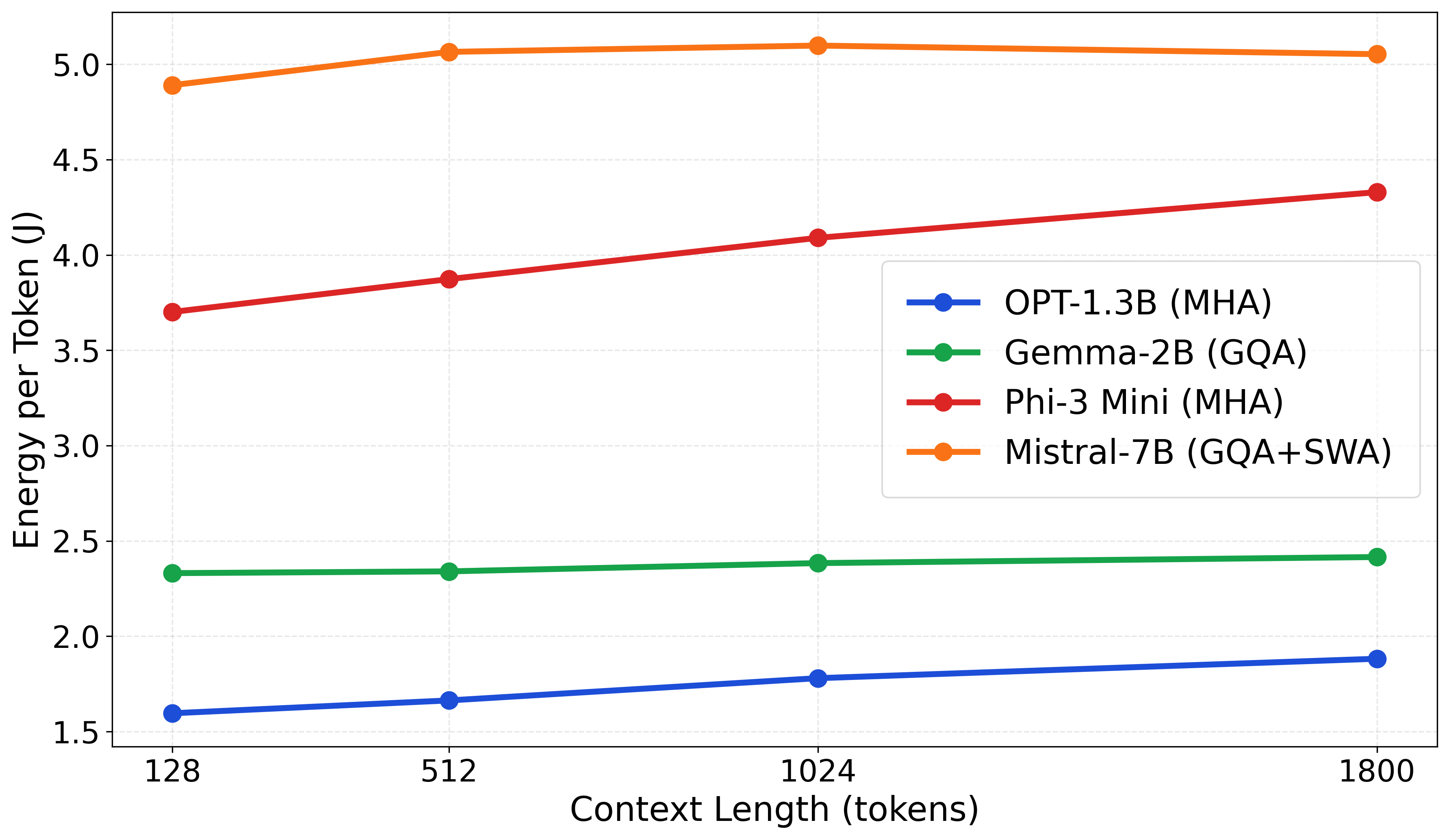}
    \caption{Decode energy per generated token across different context lengths for the evaluated models.}
    \label{fig:rq2_energy}
\end{figure}
\subsection{Effect of Context Length on Decode Energy and Latency}

Table~\ref{tab:energy_growth} and Fig.~\ref{fig:rq2_energy} show how decode energy per generated token changes with context length. Two distinct patterns emerge across the evaluated attention mechanisms. The MHA models exhibit a clear increase in energy per token as context length grows. For OPT-1.3B, energy rises from 1.5960 J at 128 tokens to 1.8820 J at 1800 tokens, corresponding to an increase of 17.92\%. Similarly, Phi-3 Mini increases from 3.7009 J to 4.3292 J, representing a 16.98\% increase. Although the increase is not perfectly linear, both models show a consistent sensitivity to context length.

In contrast, the models using more compact key-value representations exhibit substantially flatter energy profiles. Gemma-2B increases from 2.3310 J to 2.4155 J, a change of 3.62\%, while Mistral-7B varies only slightly across the evaluated context lengths, with an overall increase of 3.32\% between 128 and 1800 tokens. These results indicate that GQA reduces the additional decode-energy cost associated with longer contexts, while the combination of GQA and SWA further limits context-dependent growth. Because Mistral combines both mechanisms, its behavior should be interpreted as the joint effect of key-value head sharing and bounded attention rather than GQA alone.

The observed pattern can be explained by differences in how the models access the KV cache during decoding. In MHA models, each generated token attends over the full set of cached keys and values, increasing attention-related computation and memory access as the context expands. GQA reduces this cost by sharing key-value heads across multiple query heads, whereas SWA bounds attention to a fixed local window. The consistency of the trend across the two MHA models and the two more efficient attention designs suggests that attention architecture is an important factor governing decode-energy scaling, although model-specific implementation differences may also contribute.

\begin{figure}[t]
    \centering
    \includegraphics[width=\linewidth]{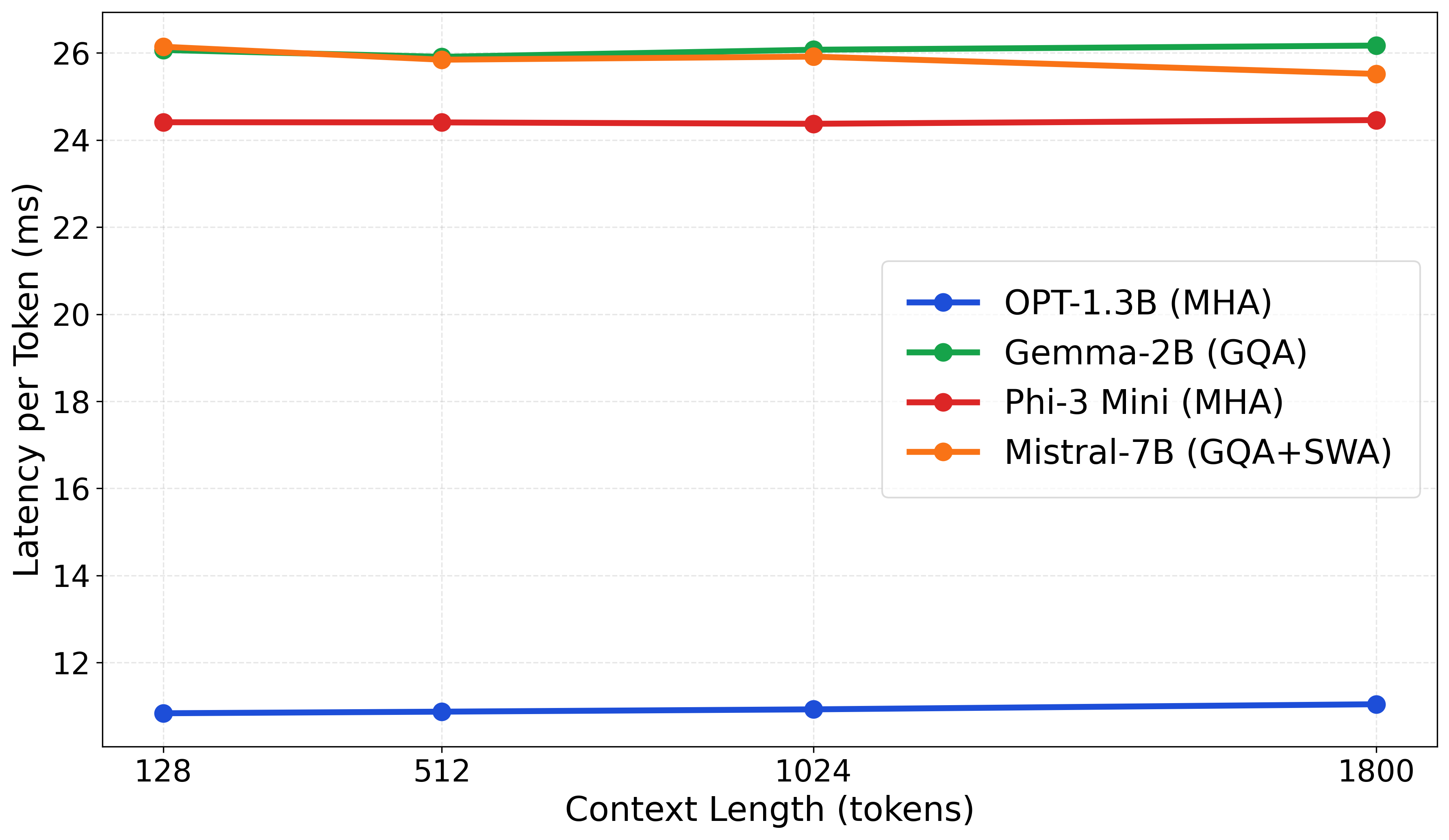}
    \caption{Latency per generated token across different context lengths for the evaluated models.}
    \label{fig:rq2_latency}
\end{figure}

As shown in Fig. \ref{fig:rq2_latency}, latency per token remains comparatively stable across context lengths. OPT-1.3B varies from 10.84 ms to 11.04 ms, Phi-3 Mini from 24.41 ms to 24.46 ms, Gemma-2B from 25.91 ms to 26.17 ms, and Mistral-7B from 25.52 ms to 26.14 ms. Since energy increases for the MHA models without a comparable increase in latency, the additional energy is consistent with higher average GPU power or hardware activity during each decoding step rather than longer token-generation time. This result also shows that latency alone may not reveal the energy cost introduced by longer contexts.

\begin{tcolorbox}[colback=gray!10,colframe=black,title=\textbf{Key Finding}]
Decode energy scaling is primarily governed by the attention mechanism rather than model size. 
MHA models exhibit a 17--18\% increase in energy as context length grows from 128 to 1800 tokens, 
whereas GQA-based models increase by only 3--4\%. In contrast, latency remains nearly constant 
across all context lengths, indicating that longer contexts primarily increase GPU energy rather 
than execution time.
\end{tcolorbox}

\begin{table}[]
\centering
\caption{Decode Energy Growth Across Context Lengths}
\label{tab:energy_growth}
\begin{tabular}{lcccc}
\hline
\textbf{Model} & \textbf{Attention} & \textbf{128→512} & \textbf{128→1024} & \textbf{128→1800} \\
\hline\hline
OPT-1.3B   & MHA     & +4.22\% & +11.53\% & +17.92\% \\
Phi-3 Mini & MHA     & +4.66\% & +10.50\% & +16.98\% \\
Gemma-2B   & GQA     & +0.40\% & +2.26\%  & +3.62\%  \\
Mistral-7B & GQA+SWA & +3.57\% & +4.24\%  & +3.32\%  \\
\hline
\end{tabular}
\end{table}

\subsection{Interplay Between Model Size and Attention Architecture}

The previous subsection showed that attention mechanism strongly influences how decode energy scales with context length. However, comparing the absolute energy and latency values across models reveals a complementary trend. While attention architecture governs the \emph{growth} of decode energy, the \emph{baseline} energy required to generate each token is primarily determined by model size.

At context length 128, OPT-1.3B exhibits the lowest energy consumption at 1.5960 J per token, followed by Gemma-2B (2.3310 J), Phi-3 Mini (3.7009 J), and Mistral-7B (4.8905 J). This ordering closely follows model size rather than attention mechanism. Although Mistral-7B employs the most efficient attention architecture among the evaluated models, it also contains the largest number of parameters, resulting in the highest absolute energy consumption. Conversely, OPT-1.3B uses the least efficient attention mechanism in terms of scaling, yet it consumes the least energy per generated token because of its substantially smaller model size. These results indicate that model size determines the baseline computational cost of decoding, whereas attention mechanism determines how this cost evolves as context length increases.

A similar trend is observed for latency. OPT-1.3B generates tokens in approximately 10.84 ms, while Phi-3 Mini, Gemma-2B, and Mistral-7B require approximately 24--26 ms per token. Interestingly, Gemma-2B and Mistral-7B exhibit nearly identical latency despite Mistral containing more than three times as many parameters. This suggests that the computational savings provided by Sliding Window Attention partially compensate for the additional computation introduced by the larger model, allowing Mistral to achieve latency comparable to a significantly smaller model.

Together, these results demonstrate that model size and attention architecture optimize different aspects of inference efficiency. Model size primarily determines the baseline energy and latency of token generation, whereas attention mechanism determines how efficiently the model scales to longer contexts. Consequently, selecting an LLM for deployment requires balancing these complementary factors according to the target application. For short-context or latency-sensitive workloads, smaller models provide lower absolute energy consumption and faster generation. In contrast, applications involving long-context generation benefit from architectures employing GQA or GQA with SWA, as they substantially reduce the additional energy introduced by increasing context length.

\begin{tcolorbox}[colback=gray!10,colframe=black,title=\textbf{Takeaway}]
Decode-phase energy consists of two complementary components: a baseline cost determined primarily by model size and a scaling cost determined by the attention mechanism. Smaller models minimize absolute energy and latency, whereas efficient attention mechanisms minimize the additional energy incurred by long-context decoding.
\end{tcolorbox}

\begin{table}[t]
\centering
\caption{Within-Sequence Energy Drift — OPT-1.3B}
\label{tab:rq3_opt}
\begin{tabular}{|c|c|c|c|}
\hline
\textbf{Context} & \textbf{E at pos 50 (J)} & \textbf{E at pos 950 (J)} & \textbf{Drift} \\
\hline
128  & 1.5973 & 1.7514 & +9.64\% \\
512  & 1.7116 & 1.8310 & +6.97\% \\
1024 & 1.8384 & 1.8901 & +2.81\% \\
\hline
\end{tabular}
\end{table}

\begin{table}[t]
\centering
\caption{Within-Sequence Energy Drift — Phi-3 Mini}
\label{tab:rq3_phi3}
\begin{tabular}{|c|c|c|c|}
\hline
\textbf{Context} & \textbf{E at pos 50 (J)} & \textbf{E at pos 950 (J)} & \textbf{Drift} \\
\hline
128  & 3.9450 & 4.1586 & +5.41\% \\
512  & 3.6976 & 3.9354 & +6.43\% \\
1024 & 4.1330 & 4.3950 & +6.34\% \\
\hline
\end{tabular}
\end{table}

\begin{table}[t]
\centering
\caption{Within-Sequence Energy Drift — Gemma-2B}
\label{tab:rq3_gemma}
\begin{tabular}{|c|c|c|c|}
\hline
\textbf{Context} & \textbf{E at pos 50 (J)} & \textbf{E at pos 950 (J)} & \textbf{Drift} \\
\hline
128  & 3.8753 & 3.9573 & +2.12\% \\
512  & 3.8075 & 3.8451 & +0.99\% \\
1024 & 4.1294 & 4.1164 & -0.32\% \\
\hline
\end{tabular}
\end{table}

\begin{table}[t]
\centering
\caption{Within-Sequence Energy Drift — Mistral-7B}
\label{tab:rq3_mistral}
\begin{tabular}{|c|c|c|c|}
\hline
\textbf{Context} & \textbf{E at pos 50 (J)} & \textbf{E at pos 950 (J)} & \textbf{Drift} \\
\hline
128  & 4.0027 & 4.0435 & +1.02\% \\
512  & 4.1296 & 4.1013 & -0.69\% \\
1024 & 4.1779 & 4.1431 & -0.83\% \\
\hline
\end{tabular}
\end{table}
\begin{figure*}[t]
    \centering
    \includegraphics[width=\textwidth]{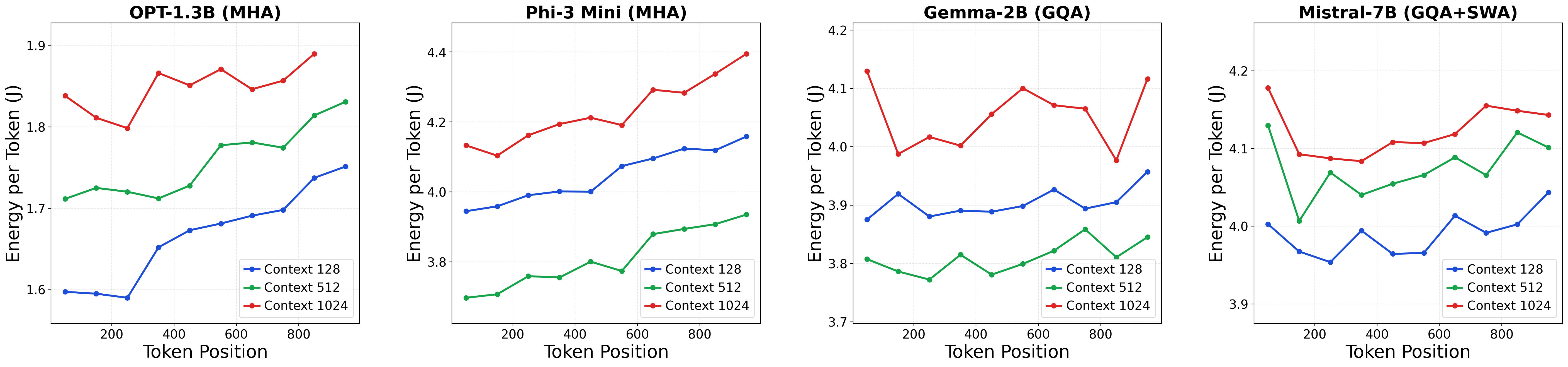}
    \caption{Energy per token vs token position across all four models and three context lengths.}
    \label{fig:rq3_energy}
\end{figure*}

\begin{figure*}[t]
    \centering
    \includegraphics[width=\textwidth]{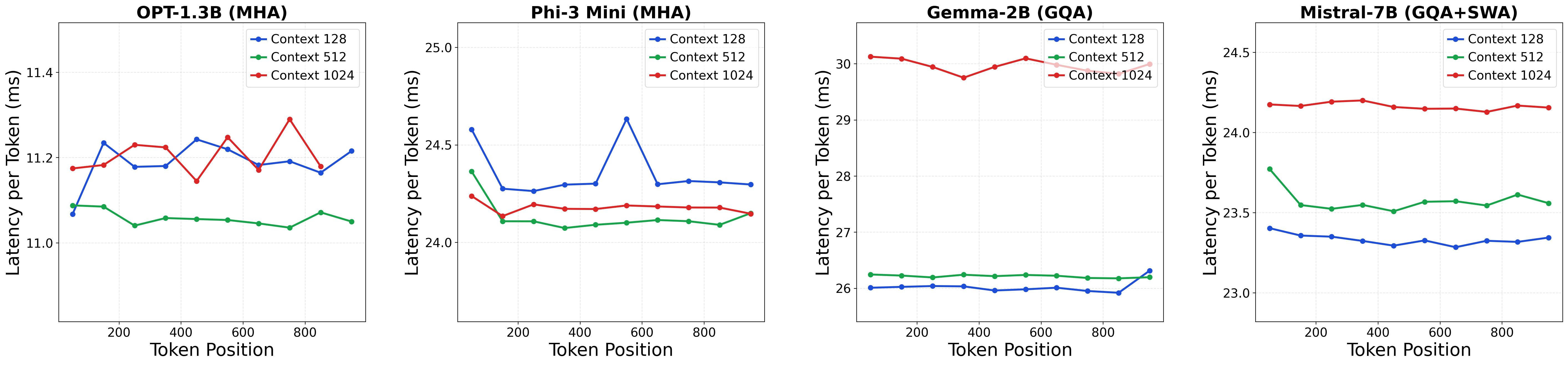}
    \caption{Latency per token vs token position across all four models and three context lengths.}
    \label{fig:rq3_latency}
\end{figure*}

\subsection{Impact of Token Position on Decode Energy}

To investigate how decode energy evolves during autoregressive generation, we measured the energy consumed every 100 generated tokens, creating a within-sequence energy profile for each model. The experiment was conducted at context lengths of 128, 512, and 1024 tokens, with each configuration repeated 10 times. Fig.~\ref{fig:rq3_energy} and Fig.~\ref{fig:rq3_latency} illustrate the energy and latency profiles across token positions, while Tables~\ref{tab:rq3_opt}--\ref{tab:rq3_mistral} summarize the within-sequence energy drift, defined as the percentage change between the first measurement (token position 50) and the last measurement (token position 950).

The two models employing Multi-Head Attention exhibit a clear increase in decode energy as generation progresses. For OPT-1.3B, energy drift ranges from 9.64\% at context length 128 to 2.81\% at context length 1024. The smaller drift at longer starting contexts suggests that each newly generated token contributes a proportionally smaller increase to the already large KV cache. Phi-3 Mini shows a similar trend, with energy drift remaining consistently between 5.41\% and 6.43\% across all three context lengths. Although the magnitude of the drift differs between the two models, both demonstrate that decode energy gradually increases as the KV cache grows during generation. This behavior is expected because each decoding step attends over the entire KV cache, increasing attention-related computation and memory access as more tokens are generated.

In contrast, the models employing Grouped Query Attention exhibit substantially flatter within-sequence energy profiles. Gemma-2B shows energy drift of only 2.12\%, 0.99\%, and $-0.32\%$ across the three context lengths, indicating that decode energy remains nearly constant throughout generation. By reducing the number of key-value heads through head sharing, GQA substantially limits the additional computation introduced as the KV cache grows. Mistral-7B exhibits the flattest energy profile among all evaluated models, with drift values of +1.02\%, $-0.69\%$, and $-0.83\%$, all of which fall within measurement variability. Unlike Gemma, Mistral combines GQA with SWA, limiting each decoding step to a fixed local attention window. As a result, the amount of attention computation remains nearly constant regardless of the number of generated tokens, effectively eliminating within-sequence energy drift.

Overall, the results reveal three distinct energy behaviors corresponding to the evaluated attention mechanisms. MHA models exhibit measurable within-sequence energy drift as the KV cache expands, GQA substantially reduces this drift, and the combination of GQA with SWA effectively eliminates it. These findings complement the results of the previous subsection by showing that attention architecture governs not only how decode energy scales with the initial context length, but also how it evolves throughout the generation process. Consequently, efficient attention mechanisms improve energy efficiency both across different workloads and within a single inference request.

\begin{tcolorbox}[colback=gray!10,colframe=black,title=\textbf{Takeaway}]
Attention architecture influences decode-energy behavior across both context lengths and token generation. MHA models exhibit increasing energy as the KV cache expands during token generation, GQA substantially reduces this effect, and GQA combined with SWA maintains nearly constant energy throughout the generation process.
\end{tcolorbox}
\begin{figure*}[t]
    \centering
    \includegraphics[width=\textwidth]{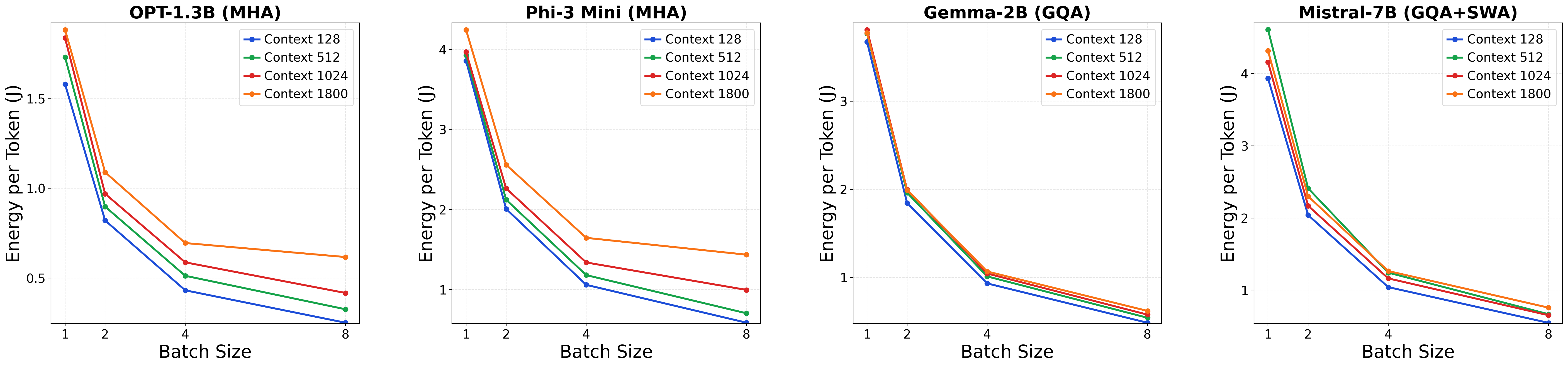}
    \caption{Energy per token vs batch size across all four models and four context lengths.}
    \label{fig:rq4_energy}
\end{figure*}

\begin{figure*}[t]
    \centering
    \includegraphics[width=\textwidth]{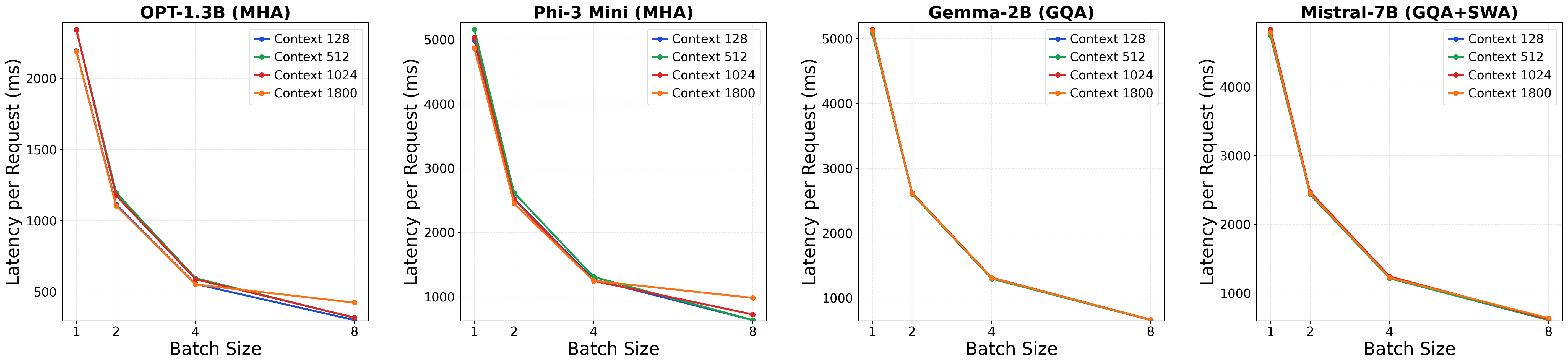}
    \caption{Latency per request vs batch size across all four models and four context lengths.}
    \label{fig:rq4_latency}
\end{figure*}

\begin{table}[t]
\centering
\caption{Impact of Batching on Energy per Token OPT-1.3B}
\label{tab:rq4_opt}
\begin{tabular}{|l|c|c|c|c|c|}
\hline
\textbf{Context} & \textbf{Batch 1} & \textbf{Batch 2} & \textbf{Batch 4} & \textbf{Batch 8} & \textbf{Reduction} \\
\hline
128  & 1.5806 & 0.8209 & 0.4315 & 0.2508 & -84.13\% \\
512  & 1.7316 & 0.8981 & 0.5125 & 0.3258 & -81.18\% \\
1024 & 1.8379 & 0.9695 & 0.5874 & 0.4167 & -77.33\% \\
1800 & 1.8839 & 1.0898 & 0.6950 & 0.6169 & -67.25\% \\
\hline
\end{tabular}
\end{table}

\begin{table}[t]
\centering
\caption{Impact of Batching on Energy per Token Phi-3 Mini}
\label{tab:rq4_phi3}
\begin{tabular}{|l|c|c|c|c|c|}
\hline
\textbf{Context} & \textbf{Batch 1} & \textbf{Batch 2} & \textbf{Batch 4} & \textbf{Batch 8} & \textbf{Reduction} \\
\hline
128  & 3.8601 & 2.0081 & 1.0591 & 0.5861 & -84.82\% \\
512  & 3.9305 & 2.1231 & 1.1809 & 0.7036 & -82.10\% \\
1024 & 3.9699 & 2.2647 & 1.3387 & 0.9959 & -74.91\% \\
1800 & 4.2470 & 2.5602 & 1.6466 & 1.4361 & -66.18\% \\
\hline
\end{tabular}
\end{table}

\begin{table}[t]
\centering
\caption{Impact of Batching on Energy per Token Gemma-2B}
\label{tab:rq4_gemma}
\begin{tabular}{|l|c|c|c|c|c|}
\hline
\textbf{Context} & \textbf{Batch 1} & \textbf{Batch 2} & \textbf{Batch 4} & \textbf{Batch 8} & \textbf{Reduction} \\
\hline
128  & 3.6735 & 1.8455 & 0.9334 & 0.4872 & -86.74\% \\
512  & 3.7679 & 1.9621 & 1.0129 & 0.5439 & -85.56\% \\
1024 & 3.8109 & 1.9965 & 1.0457 & 0.5794 & -84.80\% \\
1800 & 3.7751 & 1.9916 & 1.0686 & 0.6217 & -83.54\% \\
\hline
\end{tabular}
\end{table}

\begin{table}[t]
\centering
\caption{Impact of Batching on Energy per Token Mistral-7B}
\label{tab:rq4_mistral}
\begin{tabular}{|l|c|c|c|c|c|}
\hline
\textbf{Context} & \textbf{Batch 1} & \textbf{Batch 2} & \textbf{Batch 4} & \textbf{Batch 8} & \textbf{Reduction} \\
\hline
128  & 3.9341 & 2.0400 & 1.0421 & 0.5470 & -86.10\% \\
512  & 4.6103 & 2.4120 & 1.2428 & 0.6667 & -85.54\% \\
1024 & 4.1607 & 2.1694 & 1.2649 & 0.6524 & -84.32\% \\
1800 & 4.3162 & 2.3015 & 1.1630 & 0.7576 & -82.45\% \\
\hline
\end{tabular}
\end{table}

\subsection{Impact of Batching on Energy and Latency Efficiency}

To evaluate the impact of request batching on inference efficiency, we measured energy per generated token and latency per request using batch sizes of 1, 2, 4, and 8 across all four context lengths and models. Fig.~\ref{fig:rq4_energy} and Fig.~\ref{fig:rq4_latency} present the corresponding energy and latency trends, while Tables~\ref{tab:rq4_opt}--\ref{tab:rq4_mistral} summarize the percentage reduction in energy per token when increasing the batch size from 1 to 8.

Across all evaluated models, batching substantially reduces energy per generated token. Increasing the batch size from 1 to 8 decreases energy consumption by more than 80\% in nearly all configurations. The improvement is observed consistently for both MHA- and GQA-based models, demonstrating that batching is an effective optimization independent of the underlying attention mechanism. However, the magnitude of the improvement differs across architectures. For the MHA models, the energy reduction decreases from approximately 84\% at context length 128 to approximately 67\% at context length 1800, indicating that batching becomes less effective as context length increases. In contrast, Gemma-2B and Mistral-7B maintain reductions above 82\% across all evaluated contexts, showing little degradation in batching efficiency.

The difference between the two attention mechanisms can be explained by the growth of the KV cache. In MHA models, longer contexts require substantially more memory to store and access the KV cache for every request. As the batch size increases, this additional memory demand limits GPU parallelism and reduces the efficiency gains obtained through batching. GQA alleviates this bottleneck by reducing the number of key-value heads, while the combination of GQA and Sliding Window Attention further limits the amount of attention computation required for each generated token. Consequently, GQA-based models maintain high batching efficiency even under long-context workloads.

A similar trend is observed for latency. Increasing the batch size consistently reduces latency per request across all models, with reductions of approximately 86--87\% at context length 128. Unlike many optimization techniques that improve energy efficiency at the expense of execution time, batching simultaneously improves both metrics by increasing GPU utilization. With a single request, a large fraction of the GPU's computational resources remain idle. Processing multiple requests concurrently allows these resources to be shared across requests, reducing both the energy required to generate each token and the latency associated with each request.

Overall, these results identify batching as the most effective deployment-level optimization evaluated in this study. Unlike architectural modifications such as GQA or SWA, batching requires no changes to the underlying model and can be readily applied to existing inference systems. Although attention architecture influences the extent of the batching benefit under long-context workloads, batching consistently provides substantial improvements in both energy efficiency and latency across all evaluated models.

\begin{tcolorbox}[colback=gray!10,colframe=black,title=\textbf{Takeaway}]
Batching is the most effective deployment-level optimization evaluated in this study. It consistently improves both energy efficiency and latency across all models, while GQA and GQA combined with SWA preserve these benefits more effectively than MHA under long-context workloads.
\end{tcolorbox}
\section{Conclusion}
This study presents a systematic empirical characterization of decode-phase energy consumption across four open-source language models with fundamentally different attention mechanisms, evaluated under identical hardware and measurement conditions. Our results show that attention mechanism design is a primary factor determining how decode energy scales with context length. MHA models exhibit significantly steeper energy growth than GQA and GQA+SWA models as context increases. In terms of absolute energy consumption, model size is the dominant factor, as smaller models consume less energy per token regardless of their attention mechanism. Regarding within-sequence energy drift, MHA models show measurable energy increases during generation as the Key-Value cache grows, while GQA models remain nearly flat and Sliding Window Attention eliminates drift entirely. Finally, batching consistently and dramatically reduces both energy per token and latency per request across all models and context lengths, making it the most impactful optimization available for deployed inference systems. These findings help researchers and practitioners better understand the trade-offs between attention architecture, model size, and inference configuration, enabling more informed and energy-aware decisions when selecting and deploying large language models in production environments.

\bibliographystyle{IEEEtran}
\bibliography{biliography}
\end{document}